\documentclass[runningheads]{llncs}
\usepackage[T1]{fontenc}

\usepackage{amsmath}
\usepackage{amsfonts}
\usepackage{algorithm}
\usepackage{algpseudocode}
\usepackage{multicol}
\usepackage{graphicx}
\usepackage{multirow}
\usepackage{subcaption}

\usepackage{algpseudocode}
\usepackage{stmaryrd}

\usepackage{breakcites}
\usepackage{booktabs}
\usepackage{hyperref}
\usepackage{color}

\hypersetup{
    colorlinks,
    linkcolor={blue},
    citecolor={blue},
    urlcolor={blue}
}

\usepackage{cleveref}

\begin{document}

\title{International Transfer of Stochastic Cortical Self-Reconstruction}
\titlerunning{International Transfer of Stochastic Cortical Self-Reconstruction}

\author{%
  Fabian Bongratz\inst{1,2*}\and
  Zhizheng Zhuo\inst{5}\and
  Chao Zhang\inst{5}\and
  Yaou Liu\inst{5}\and
  Dennis M. Hedderich\inst{4}\and
  Christian Wachinger\inst{1,2,3}
}
\authorrunning{F. Bongratz, Z. Zhuo, C. Zhang, Y. Liu, D. M. Hedderich, C. Wachinger}

\institute{Lab for AI in Medical Imaging, Institute for Diagnostic and Interventional Radiology, School of Medicine and Health, TUM Klinikum, Technical University of Munich (TUM), Munich, 81675, Germany 
\and
Munich Center for Machine Learning (MCML), Munich, Germany
\and
Munich Data Science Institute (MDSI), Technical University of Munich, Munich, Germany
\and
Department of Neuroradiology, School of Medicine and Health, TUM Klinikum, Technical University of Munich (TUM), Munich, 81675, Germany
\and
Department of Radiology, Beijing Tiantan Hospital, Capital Medical University, No. 119, West Southern 4th Ring Road, Fengtai District, Beijing, 100070, China. 
\\ *Correspondence: \email{fabi.bongratz@tum.de}
}

\maketitle

\begin{abstract}
Stochastic cortical self-reconstruction (SCSR) enables personalized mapping of gray matter atrophy, a hallmark of neurodegenerative disorders such as Alzheimer’s disease (AD), onto high-resolution cortical surfaces. Unlike conventional normative modeling approaches, which typically operate at a coarse regional level and remain inherently constrained by the covariates included during training, SCSR estimates an individualized healthy reference directly from the observed cortical thickness at the vertex level. This allows the detection of subtle, subject-specific deviations from healthy cortical shape.
In this work, we investigate the generalization and transferability of SCSR, originally trained on UK Biobank (UKB) data, to an independent Chinese population dataset. Specifically, we evaluate the ability of SCSR-derived Z-scores to discriminate between healthy scans, individuals with mild cognitive impairment (MCI), and patients with AD, while also assessing model robustness across the lifespan. We compare four training strategies: direct application of the UKB-trained model, fine-tuning on Chinese data, training from scratch, and joint training on UKB and Chinese cohorts. As reconstruction backbones, we consider both a multilayer perceptron (MLP) and a Spherical UNet (SUNet).
Our results demonstrate that SCSR provides robust detection of cortical atrophy in the Chinese population across all evaluated models. The highest discriminative performance was achieved by the fine-tuned SUNet model (average pairwise AUC = 0.848), followed closely by the UKB-trained SUNet. Moreover, reconstruction errors remained low across the lifespan, even when the training population exhibited a substantially narrower age distribution, indicating strong cross-population transferability.

\end{abstract}
\section{Introduction}
Detecting and visualizing individual gray matter loss has large potential for improving the diagnosis and clinical care for psychiatric and neurodegenerative disorders, which are often associated with subtle and spatially heterogeneous patterns of structural brain changes~\cite{risacher2013neuroimaging}. In particular, cortical thickness serves as a sensitive biomarker for distinguishing different dementia types~\cite{du2007different} and predicting disease progression, such as the transition from mild cognitive impairment (MCI) to manifest Alzheimer's disease (AD)~\cite{Dickerson2013prediction}. However, identifying and mapping cortical thickness changes at the individual level remains difficult due to substantial inter-subject variability in brain anatomy~\cite{frangou2022cortical} and the influence of scanner, acquisition, and software on the measurements obtained~\cite{wachinger2021detect,SchmitzKoep2025fscat}. In practice, this variability affects the transfer of developed models and techniques across international, multi-site datasets that differ in scanner hardware, acquisition protocols, and population demographics.

Reference, or normative, models have emerged as a way to estimate the expected range of brain measurements, such as coarse volume and surface descriptors, as a function of covariates, such as age and sex~\cite{bethlehem2022brain,rutherford2022normative}. By comparing the actual measurement against the expected range, individual deviations are quantified as centiles or Z-scores. Such models have been applied across a wide range of neurological, developmental, and psychiatric conditions~\cite{marquand2019conceptualizing,Thalhammer2025heterogeneous,Zhuo2025charting}. Nevertheless, this modeling paradigm remains inherently limited to the considered covariates, and the generated reference is therefore not tailored to an individual's unique brain morphology. Moreover, the covariate-based modeling approach does not scale easily to thousands of cortical thickness measurements across the cortical ribbon, as obtained from surface-based neuroimaging tools like FreeSurfer~\cite{fischl2012freesurfer}.

Stochastic Cortical Self-Reconstruction (SCSR)~\cite{Wachinger2026SCSR} was proposed as a reference model that sidesteps the explicit covariate-based formulation of classical normative models. Instead of mapping demographic and site-describing variables to an expected cortical phenotype, SCSR repeatedly reconstructs an individual's full cortical morphology from a stochastically sampled subset of measurements, using a neural network trained exclusively on healthy individuals. A personalized reference distribution is obtained via repeated sampling of the subset, from which standardized Z-scores are computed.

In this work, we study the transfer of Stochastic Cortical Self-Reconstruction (SCSR)~\cite{Wachinger2026SCSR}, originally trained and developed on UK Biobank imaging data~\cite{littlejohns2020uk}, to a Chinese population dataset.
Unlike UKB, which covers only ages 45 to 82, this cohort spans ages 4 to 85, including children and adolescents.
We evaluate four strategies for applying SCSR to this new population: (1) direct application of the UKB-trained model without any adaptation, (2) fine-tuning the UKB-trained model on the Chinese cohort, (3) training a model from scratch using only the Chinese cohort, and (4) jointly training on UKB and the Chinese cohort. We repeat all four strategies for both the MLP and Spherical UNet (SUNet) implementations of SCSR, resulting in eight experimental configurations. We quantify each configuration's ability to detect atrophy in AD and MCI patients from the same population via pairwise AUC on SCSR-derived Z-scores. Our results characterize the extent to which population-specific adaptation contributes on top of a large, out-of-population pretraining source and whether this benefit depends on the architecture choice.

\section{Related Work}

Normative models of regional cortical thickness trajectories across the lifespan have been developed using generalized additive models of location, scale, and shape (GAMLSS)~\cite{bethlehem2022brain,rigby2005generalized}. Analogous to pediatric growth charts~\cite{Borghi2005growthcharts}, these models provide age- and sex-specific reference distributions against which an individual's brain morphology can be compared. By jointly modeling the mean, variance, skewness, and kurtosis, GAMLSS allows the entire distribution of cortical morphometry to change nonlinearly across the lifespan. Alternative approaches based on Bayesian regression and other flexible predictive frameworks have likewise been proposed, for example, within the PCNToolkit ecosystem~\cite{rutherford2022normative}. Normative modeling has been applied to characterize inter-individual heterogeneity in a range of neurological and psychiatric conditions, including schizophrenia~\cite{di2022cell,Volkmer2026schizo}, preterm birth~\cite{Thalhammer2025heterogeneous}, and autism spectrum disorder~\cite{Zabihi2019autism}, as well as to characterize variation across the whole body \cite{wachinger2025body,wachinger2026whole}; see~\cite{marquand2019conceptualizing} for a comprehensive overview. 

Deep learning for cortical surface analysis has largely focused on discriminative tasks. Common architectures operating on FreeSurfer-derived cortical meshes~\cite{fischl2012freesurfer} include spherical UNet variants with specialized graph-convolutional operators~\cite{zhao2019spherical} and transformer-based models adapted to spherical domains~\cite{Cheng2022spherical,dahan2022sit}. These methods have achieved strong performance on cortical parcellation, disease classification, and regression tasks, but they are less commonly applied to generate subject-specific normative references or model healthy anatomical variation.

The transferability of geometric deep learning representations trained on large biomedical datasets has been studied for other surface- and mesh-based tasks, such as topology-informed segmentation~\cite{Wu2026geotopo} and shape classification~\cite{Gao2021transfer}. We extend this line of work to a different class of objective: rather than transferring a representation toward a discriminative segmentation or classification target, we study the transfer of a
self-reconstruction-based reference model, whose network must recover an individualized healthy cortical thickness pattern rather than predict a fixed label.

\section{Methods}
\subsection{Data}

\noindent
\textbf{UK Biobank.}
We use the UK Biobank imaging~\cite{littlejohns2020uk} cohort of neurologically healthy subjects, split into 25,338 subjects for training and 6,334 for validation; further details are provided in~\cite{Wachinger2026SCSR}.

\vspace{\baselineskip}
\noindent
\textbf{Chinese Population Dataset.}
From a Chinese population dataset~\cite{Zhuo2025charting}, we use 640 healthy scans for training/finetuning and 160 healthy scans for validation, with a balanced age distribution across almost the entire lifespan (4--86 years). The lifespan test set comprises 139 subjects, stratified into 10-year age brackets. For the atrophy detection experiments in Alzheimer's disease (AD) and mild cognitive impairment (MCI), we use 60 subjects per diagnostic group together with 60 age-matched (46--85 years) cognitively normal (CN) controls drawn from the lifespan test set.

\vspace{\baselineskip}
\noindent
\textbf{Preprocessing.}
All scans were processed with FreeSurfer~\cite{fischl2012freesurfer}, including spherical registration of the left hemisphere to the FsAverage template and resampling to the icosahedron of order five, yielding 10,242 vertex-wise thickness values used by SCSR.

\subsection{Stochastic Cortical Self-Reconstruction}

Stochastic Cortical Self-Reconstruction (SCSR) estimates a subject-specific healthy cortical reference by repeatedly reconstructing a subject's cortical thickness map from a randomly sampled subset of its own vertices~\cite{Wachinger2026SCSR}. Let $X \in \mathbb{R}^p$ denote the $p$-dimensional vertex-wise thickness map of a healthy training subject. During training, a fraction $s$ of its vertices is stochastically sampled as predictors $X_{\text{pred}}$, and the remaining vertices serve as responses $X_{\text{res}}$, such that $X = X_{\text{pred}} \cup X_{\text{res}}$. A neural network $f_\theta$ is then optimized across all healthy training subjects to minimize the reconstruction error of the masked vertices, i.e.,
\begin{equation}
    \| X_{\text{res}} - f_\theta(X_{\text{pred}}) \|^2 .
    \label{eq:training}
\end{equation}

\begin{figure}[t]
    \centering
    \includegraphics[width=0.9\linewidth]{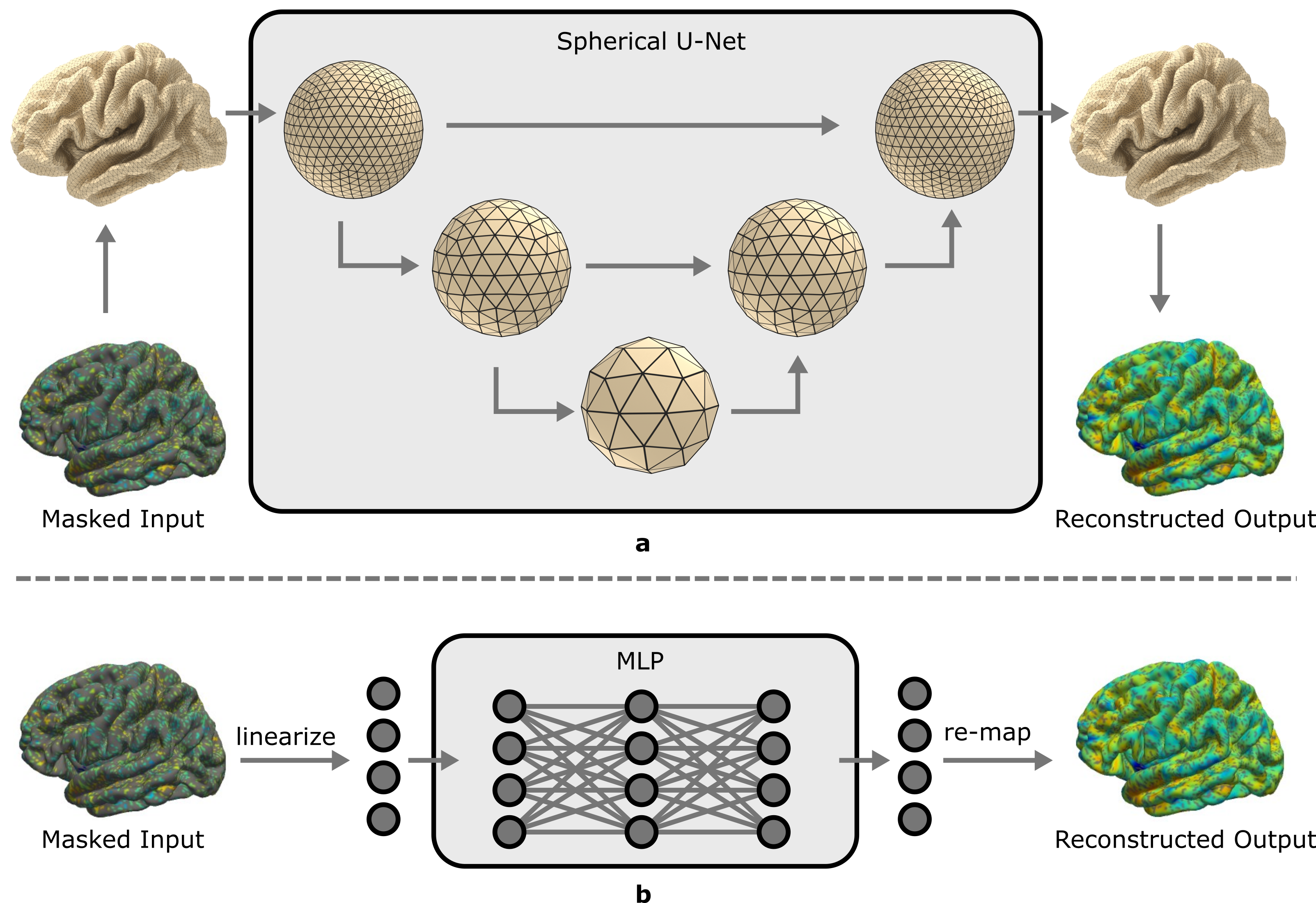}
    \caption{We study two different neural network architectures for the reconstruction in SCSR. a. Spherical UNet Model with graph convolutions optimized for icosahedral meshes. b. Multi-layer perceptron (MLP) with fully connected layers operating on linearized vertex representations.}
    \label{fig:archi}
\end{figure}

At test time, given a subject's thickness map $Y \in \mathbb{R}^p$, this stochastic sampling and prediction is repeated $m$ times: for each repetition $i = 1, \ldots, m$, a new random subset $Y_{\text{pred}}^{(i)}$ of $s\cdot p$ vertices is sampled and passed through $f_\theta$, storing the resulting prediction in the $i$-th column of a tensor $T \in \mathbb{R}^{p \times m}$ (with unsampled, i.e., unpredicted, vertex entries left as \texttt{NaN}). The subject-specific healthy reference $R \in \mathbb{R}^p$ is then obtained, vertex-wise, as the reconstruction centile $q$ of $T$ across repetitions, ignoring \texttt{NaN} entries:
\begin{equation}
    R = \operatorname{centile}(T, q) .
    \label{eq:reference}
\end{equation}
Finally, a vertex-wise Z-score quantifying the deviation from the healthy reference is computed as
\begin{equation}
    Z = \frac{Y - R}{\sigma} ,
    \label{eq:zscore}
\end{equation}
where $\sigma$ is the standard deviation of the reconstruction residuals $Y - R$, estimated on the validation set. We follow the original SCSR configuration throughout, with a sampling rate of $s=20\%$, $m=100$ repetitions, and a reconstruction centile of $q=0.95$.

We consider the two neural network implementations of SCSR visualized in \Cref{fig:archi}: a multilayer perceptron (MLP) operating on the vectorized, 10,242-dimensional per-hemisphere thickness map (20,055,608 parameters), and a Spherical UNet (SUNet)~\cite{zhao2019spherical} with 4 levels and an initial channel size of 32 at the first layer (doubling at each level) that performs spherical convolutions directly on the icosahedral mesh (1,669,217 parameters). The MLP is roughly twelve times larger than the SUNet, but does not encode any spatial inductive bias, treating the cortex as an unordered feature vector.
Our implementation is based on the public SCSR repository, available online at \url{https://github.com/ai-med/SCSR-core}.

\subsection{Experimental design}

We evaluate four strategies for adapting SCSR to the Chinese cohort, applied independently to both the MLP and SUNet architectures, resulting in eight configurations:
\begin{itemize}
\item \textbf{Direct application:} A model trained on UKB is applied to the Chinese test set without any further training.
\item \textbf{Fine-tuning:} The UKB-trained model is fine-tuned on the Chinese training set.
\item \textbf{Scratch:} A model is trained exclusively on the Chinese training set, using the same architecture and hyperparameters as the UKB models.
\item \textbf{Joint:} A model is trained jointly on the pooled UKB and Chinese training data.
\end{itemize}
All models use identical architectures, sampling rate, and reconstruction centile (see definitions above); only the training data composition differs across configurations.

\subsection{Evaluation}

\noindent
\textbf{Reconstruction error.}
As a direct measure of how well SCSR generalizes to the target population, independent of any particular downstream task, we measure the reconstruction error as the mean absolute residual between the observed thickness map $Y$ and the SCSR reference $R$, averaged vertex-wise over the full cortex:
\begin{equation}
    \text{MAE} = \frac{1}{p} \sum_{j=1}^{p} |Y_j - R_j| .
    \label{eq:mae}
\end{equation}
This uses the same per-vertex residual $Y - R$ that also enters the Z-score (\Cref{eq:zscore}), but expressed directly in mm rather than normalized by $\sigma$, so it reflects the raw fidelity with which SCSR reconstructs a subject's own healthy cortex. We report the reconstruction error per subject, aggregated by 10-year age brackets for the lifespan cohort.

\vspace{\baselineskip}
\noindent
\textbf{Atrophy detection.}
Beyond reconstruction fidelity, the actual downstream use case of SCSR is to detect atrophy from the resulting Z-scores. We therefore additionally quantify atrophy detection using the area under the receiver operating characteristic curve (AUC), computed on the mean Z-score within the AD region of interest (ROI), comprising the entorhinal, inferior temporal, middle temporal, inferior parietal, and fusiform cortices~\cite{schwarz2016large}. We compute the AUC for each of the three pairwise diagnostic comparisons, CN vs.\ MCI, CN vs.\ AD, and MCI vs.\ AD, and report their average as a single summary score per configuration.

\section{Results and Discussion}

\subsection{Reconstruction across age groups and populations}

\begin{figure}[t]
    \centering
    \includegraphics[width=0.95\linewidth]{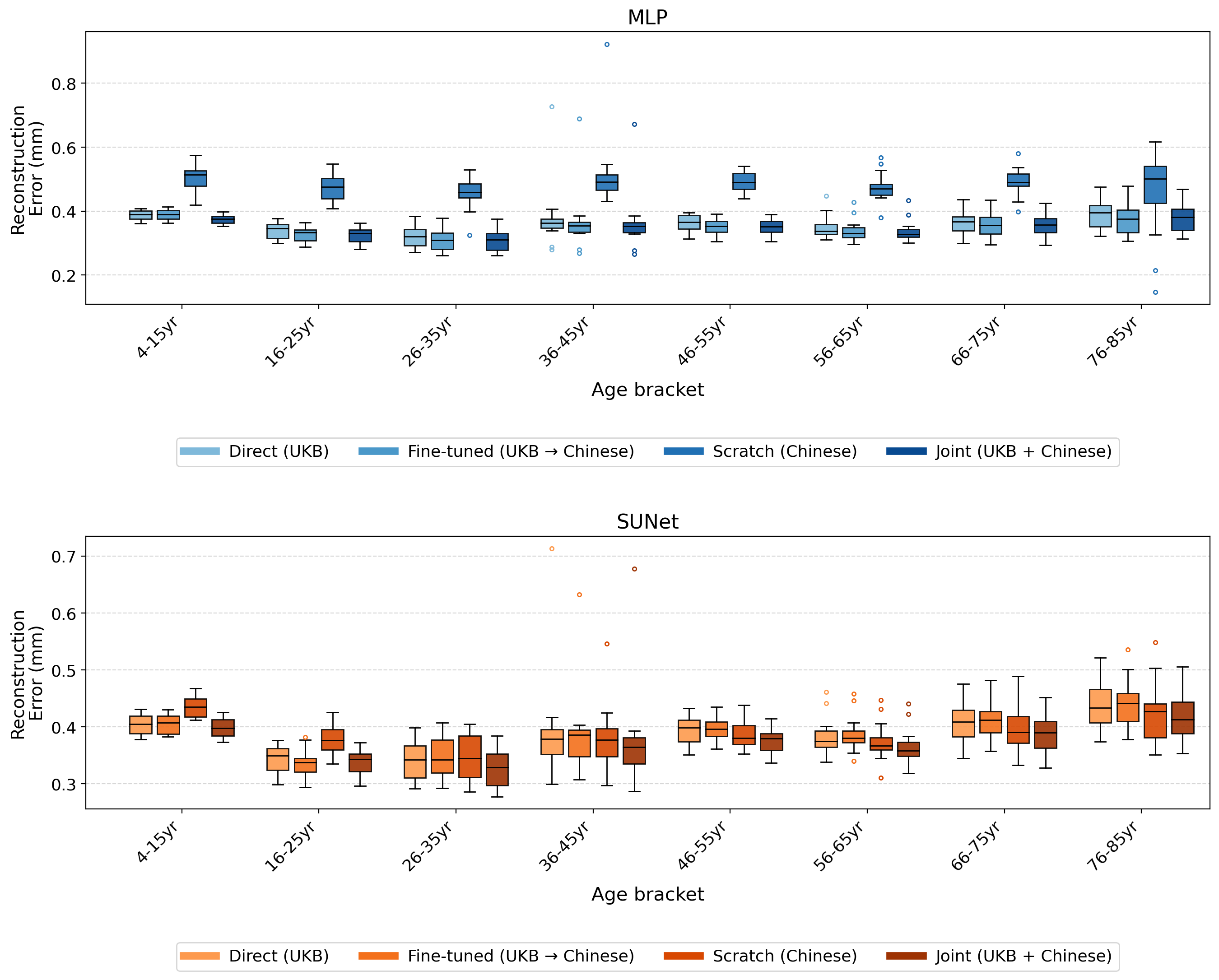}
    \caption{Distribution of the mean absolute reconstruction error per subject on the Chinese healthy lifespan cohort (4--85 years), across 10-year age brackets, for the four MLP (top) and four SUNet (bottom) configurations.}
    \label{fig:age-boxplot}
\end{figure}

\begin{table}[h!]
\setlength{\tabcolsep}{9pt}
\renewcommand\bfdefault{b}
\centering
\caption{Mean absolute reconstruction error (mm) on the Chinese lifespan test cohort (139 subjects) and the held-out UKB validation set (6,334 subjects), for all eight configurations. Best value per column in bold.}
\label{tab:mae}
\begin{tabular}{llcc}
\toprule
Backbone & Adaptation (Train Set) & Chinese & UKB \\
\midrule
\multirow{4}{*}{MLP}
 & Direct (UKB)                           & 0.359 $\pm$ 0.050 & 0.256 $\pm$ 0.025 \\
 & Fine-tuned (UKB $\rightarrow$ Chinese) & 0.349 $\pm$ 0.048 & 0.285 $\pm$ 0.023 \\
 & Scratch (Chinese)                      & 0.481 $\pm$ 0.070 & 0.454 $\pm$ 0.033 \\
 & Joint (UKB + Chinese)                  & \textbf{0.347 $\pm$ 0.047} & \textbf{0.250 $\pm$ 0.023} \\
\midrule
\multirow{4}{*}{SUNet}
 & Direct (UKB)                           & 0.385 $\pm$ 0.052 & 0.296 $\pm$ 0.020 \\
 & Fine-tuned (UKB $\rightarrow$ Chinese) & 0.386 $\pm$ 0.050 & 0.329 $\pm$ 0.021 \\
 & Scratch (Chinese)                      & 0.387 $\pm$ 0.046 & 0.348 $\pm$ 0.027 \\
 & Joint (UKB + Chinese)                  & 0.370 $\pm$ 0.049 & 0.281 $\pm$ 0.020 \\
\bottomrule
\end{tabular}
\end{table}

We first evaluate the reconstruction error across nearly the entire lifespan, using the healthy Chinese lifespan cohort. \Cref{fig:age-boxplot} shows the distribution of per-subject reconstruction error across age brackets, separately for the MLP and SUNet configurations, and \Cref{tab:mae} summarizes the corresponding cohort-level averages alongside the UKB validation results discussed below.

Averaged across the full lifespan cohort, the MLP configurations achieve the lowest reconstruction error overall, with the jointly trained (0.347 $\pm$ 0.047 mm), fine-tuned (0.349 $\pm$ 0.048 mm), and directly applied (0.359 $\pm$ 0.050 mm) MLP outperforming the SUNet configurations. The one exception is the MLP trained exclusively on the Chinese cohort (scratch), which is by far the worst configuration overall (0.481 $\pm$ 0.070 mm). We attribute this to the MLP's large parameter count (20M) and lack of a spatial inductive bias, which make it prone to overfitting when trained on only 640 Chinese subjects without the benefit of UKB pretraining. The four SUNet configurations, by contrast, are tightly clustered (0.370--0.387 mm) regardless of adaptation strategy, including SUNet trained from scratch. As for the MLP, the jointly trained SUNet model achieved, on average, the lowest reconstruction error on the Chinese data (0.370 $\pm$ 0.049 mm).

Notably, the direct application of the UKB-trained models, without any exposure to Chinese data, generalizes surprisingly well across both architectures: their reconstruction errors are comparable to those of models trained or fine-tuned on Chinese data. This holds even in the pediatric and adolescent age brackets (4--20 years), which fall entirely outside UKB's covered age range (45+) and were not specifically targeted by any adaptation strategy. Both architectures show a mild increase in reconstruction error at the youngest and oldest age brackets relative to the middle of the distribution, consistent with the increased inter-individual variability in cortical thickness reported at the extremes of the lifespan~\cite{bethlehem2022brain,frangou2022cortical,Zhuo2025charting}.

Re-evaluating all eight configurations on the held-out UKB validation set further shows that fine-tuning on the Chinese data yields a modest increase in UKB reconstruction error (MLP: 0.256 $\to$ 0.285 mm; SUNet: 0.296 $\to$ 0.329 mm). Joint training, on the other hand, improves the reconstruction error on UKB relative to the original (UKB-only) trained models (MLP: 0.250 vs.\ 0.256 mm; SUNet: 0.281 vs.\ 0.296 mm). This mirrors the behavior on the Chinese cohort, where joint training was likewise the best-performing configuration in terms of reconstruction error (see results above). The scratch-trained (Chinese-only) models generalize back to UKB the worst, as expected (0.454 $\pm$ 0.033 mm for the MLP and 0.348 $\pm$ 0.027 mm for SUNet). Nevertheless, these results confirm that the SUNet-based SCSR model is more robust to a smaller training set than the MLP-based implementation.

\begin{table}[t]
\setlength{\tabcolsep}{4.8pt}
\renewcommand\bfdefault{b}
\centering
\caption{Pairwise AUC for atrophy detection on the Chinese test cohort (60 CN, 60 MCI, 60 AD), based on mean SCSR Z-scores in the AD ROI. Best value per column in bold.}
\label{tab:auc}
\begin{tabular}{llcccc}
\toprule
Backbone & Adaptation (Train Set) & CN|MCI & CN|AD & MCI|AD & Average \\
\midrule
\multirow{4}{*}{MLP}
 & Direct (UKB)         & 0.543 & 0.841 & 0.775 & 0.720 \\
 & Fine-tuned (UKB $\rightarrow$ Chinese)           & 0.581 & 0.849 & 0.765 & 0.732 \\
 & Scratch (Chinese)     & 0.598 & 0.789 & 0.711 & 0.699 \\
 & Joint (UKB + Chinese) & 0.524 & 0.801 & 0.753 & 0.693 \\
\midrule
\multirow{4}{*}{SUNet}
 & Direct (UKB)          & 0.718 & 0.925 & 0.803 & 0.815 \\
 & Fine-tuned (UKB $\rightarrow$ Chinese)             & \textbf{0.787} & \textbf{0.938} & \textbf{0.818} & \textbf{0.848} \\
 & Scratch (Chinese)      & 0.631 & 0.863 & 0.801 & 0.765 \\
 & Joint (UKB + Chinese)  & 0.698 & 0.928 & 0.826 & 0.817 \\
\bottomrule
\end{tabular}
\end{table}

\subsection{Identifying atrophy in Alzheimer's disease and mild cognitive impairment}

We now turn to the downstream clinical task that motivates SCSR: detecting cortical atrophy in patients with AD and MCI directly from vertex-wise Z-scores, without requiring an explicitly matched normative reference cohort. \Cref{tab:auc} reports pairwise AUC scores across all eight configurations.

\begin{figure}[t]
    \centering
    \includegraphics[width=0.9\linewidth]{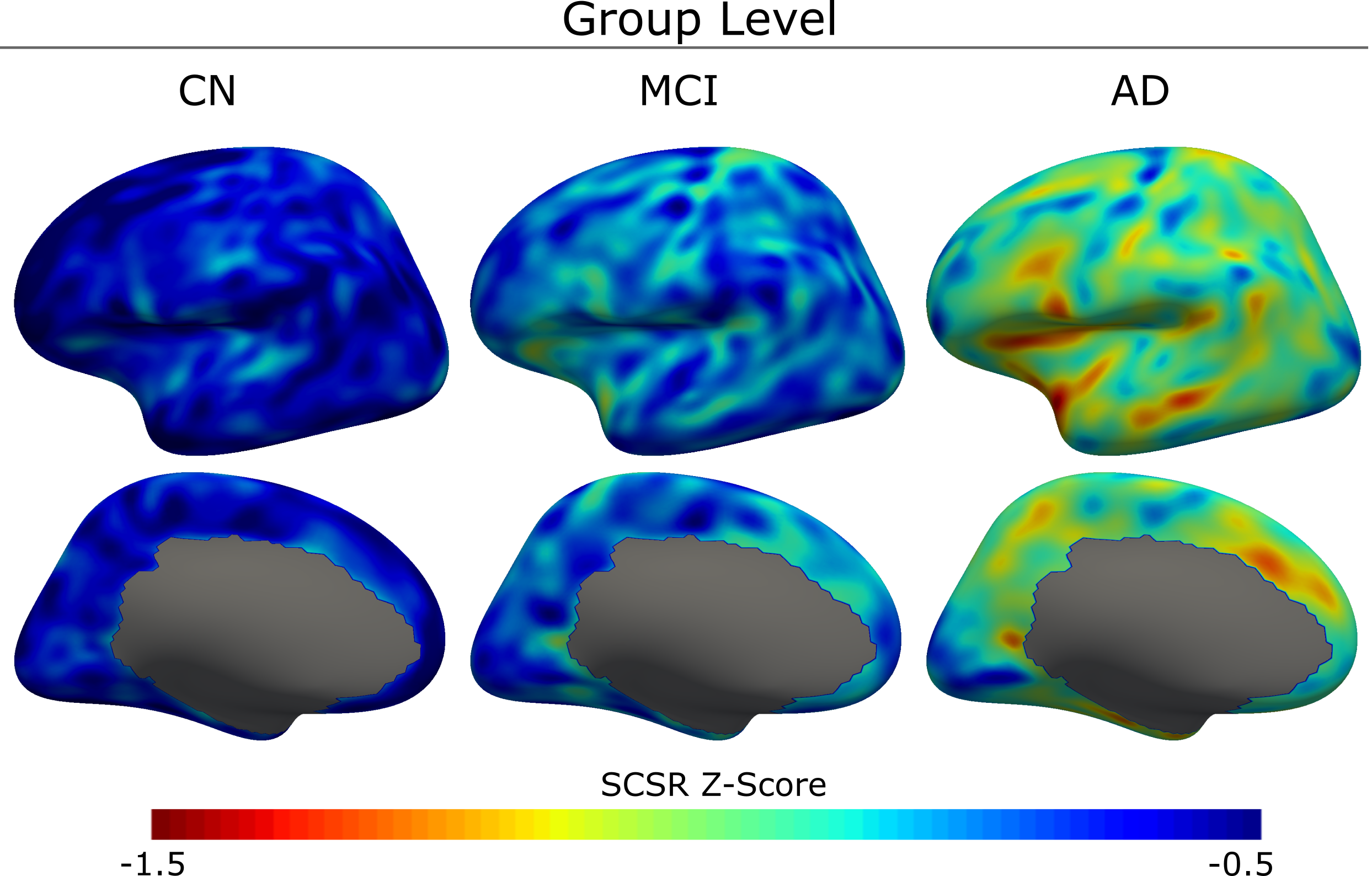}
    \caption{Vertex-wise group mean SCSR Z-scores for the three diagnostic groups: age-matched cognitively normal (CN), mild cognitive impairment (MCI), and Alzheimer's disease (AD). The visualization was created based on the finetuned SUNet model. Lower Z-scores indicate more severe atrophy. The medial region that connects the two brain hemispheres is not included in SCSR and is therefore grayed out in the visualization. Rendering was performed using PyVista.}
    \label{fig:group-surf}
\end{figure}

For atrophy detection, SUNet outperforms the matched MLP counterpart on the Chinese population data throughout all considered training configurations, by an average AUC margin ranging from 0.066 (scratch) to 0.124 (joint training). Yet, every one of the eight configurations, including the scratch-trained MLP with its markedly elevated reconstruction error (cf.\ \Cref{fig:age-boxplot}; 0.699 average AUC), remains clearly above chance, indicating that all eight models learned a healthy reference that provides a viable basis for atrophy detection, even when its absolute reconstruction fidelity is comparatively poor.

Comparing adaptation strategies, fine-tuning is the most effective choice for SUNet in this downstream task: it improves the average AUC from 0.815 (direct application) to 0.848, the best result overall, and also outperforms joint training (0.817). For the MLP, however, fine-tuning yields only a marginal improvement over direct application (0.732 vs.\ 0.720), and both training from scratch (0.699) and joint training (0.693) perform worse than simply applying the UKB-trained MLP directly. This suggests that adapting the substantially larger MLP requires more target-domain data than the 640 Chinese training subjects provide, whereas SUNet's smaller capacity and built-in spatial structure make it far more data-efficient to adapt. Still, direct out-of-population transfer is a strong baseline across both architectures and the easiest adaptation strategy to deploy in practice, requiring no additional training at all. Since joint training pools UKB and Chinese datasets without accounting for their substantial size imbalance (25K vs.\ 640 samples), a re-weighting strategy that up-weights the smaller Chinese cohort during joint training might further improve its AUC, particularly for the MLP, where joint training currently trails direct application.

\begin{figure}[t]
    \centering
    \includegraphics[width=0.9\linewidth]{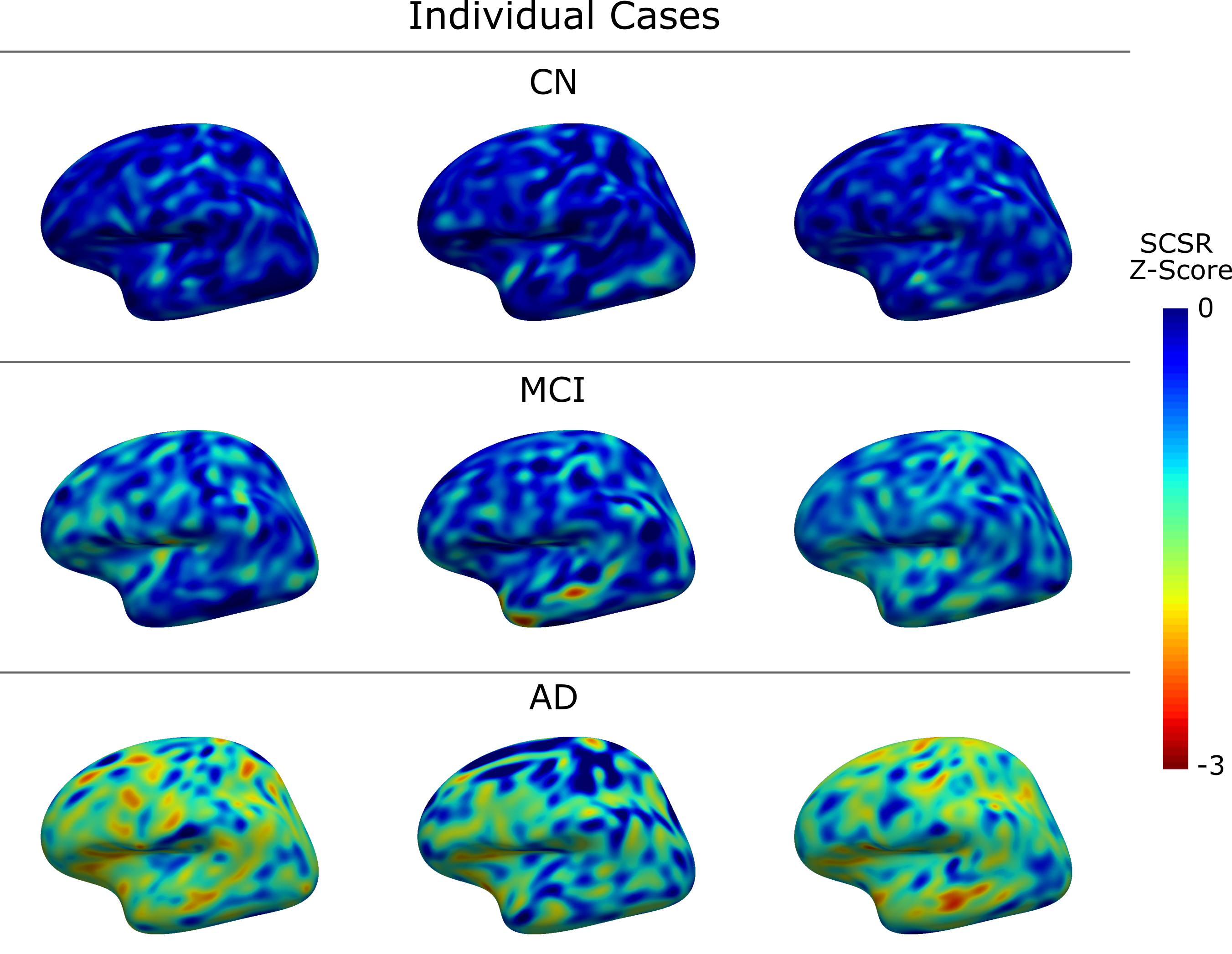}
    \caption{Individual SCSR Z-score maps computed based on three subjects from each diagnostic group (CN, MCI, AD), respectively. The visualization was created based on the finetuned SUNet model. Lower Z-scores indicate more severe atrophy. Rendering was performed using PyVista.}
    \label{fig:individual-surf}
\end{figure}

To complement the quantitative evaluation, we next investigate the spatial patterns underlying these classification results using the finetuned SUNet model. To this end, \Cref{fig:group-surf} shows SCSR outputs as vertex-wise group averages based on the 60 CN, MCI, and AD samples in the Chinese population data, respectively.
A clear, expected trend is observable, highlighting the potential of SCSR for atrophy quantification and mapping. The group-level patterns in AD are particularly pronounced in temporoparietal and medial temporal regions, as well as in frontal areas, all of which are known to be affected in the disease course~\cite{du2007different}. MCI exhibits an intermediate pattern between CN and AD, with more subtle, yet spatially consistent, cortical thinning in similar regions, reflecting the transitional nature of the disease stage.

Finally, corresponding individual Z-score maps shown in \Cref{fig:individual-surf} reveal substantial heterogeneity within diagnostic groups, highlighting considerable inter-subject variation in the spatial extent and severity of cortical thinning. This observation underscores the limitations of population-level averages for characterizing individual disease manifestations and illustrates the value of normative surface-based analysis, as provided by SCSR, for detecting and visualizing patient-specific patterns of neurodegeneration.

\section{Conclusion}

We evaluated the transfer of Stochastic Cortical Self-Reconstruction (SCSR), a neural network-based personalized normative reference model for cortical surfaces, to an independent Chinese cohort spanning a much wider age range (4--85 years) than the original, UK Biobank (UKB)-based training population. Across nearly the entire lifespan, the direct application of UKB-trained models generalizes remarkably well, with reconstruction error comparable to that of models trained or fine-tuned on Chinese data, including in pediatric and adolescent age brackets that UKB does not cover at all. The MLP consistently achieves the lowest reconstruction error, except when trained exclusively on the smaller Chinese cohort, where it appears to overfit; SUNet remains robust to the choice of adaptation strategy throughout.
For the downstream task of detecting atrophy in AD and MCI, SUNet outperformed the MLP in our experiments, with fine-tuning on the Chinese cohort achieving the best average AUC (0.848). 
This points to a practical, goal-dependent recommendation: fine-tuning is preferable when the priority is maximizing downstream diagnostic discrimination, while joint training is the better choice when the priority is retaining the best possible reconstruction accuracy across populations. Together with the strong generalization we observed even without any target-domain adaptation, we believe these results provide practical guidance for transferring SCSR to new international cohorts.

\begin{credits}
\subsubsection{\ackname} This research was partially supported by the German Research Foundation (DFG, No. 460880779).

\end{credits}

\bibliographystyle{splncs04}
\bibliography{bibliography}

\end{document}